\documentclass{article}
\usepackage{spconf,amsmath,graphicx}
\usepackage{framed,multirow}
\usepackage{amssymb}
\usepackage[dvipsnames]{xcolor}
\usepackage{soul}
\usepackage{booktabs}

\title{Stereo-Matching Knowledge Distilled Monocular Depth Estimation filtered by Multiple Disparity Consistency}
\twoauthors
 {Woonghyun Ka\sthanks{The authors are equally contributed.}}
	{Hyndai Motor Company\\
     Seoul, South Korea}
 {Jae Young Lee$^{*}$ \quad Jaehyun Choi \quad Junmo Kim}
	{KAIST, Electrical Engineering\\
	Daejeon, South Korea}
\begin{document}
%
\maketitle
\begin{abstract}
In stereo-matching knowledge distillation methods of the self-supervised monocular depth estimation, the stereo-matching network's knowledge is distilled into a monocular depth network through pseudo-depth maps.
In these methods, the learning-based stereo-confidence network is generally utilized to identify errors in the pseudo-depth maps to prevent transferring the errors.
However, the learning-based stereo-confidence networks should be trained with ground truth (GT), which is not feasible in a self-supervised setting.
In this paper, we propose a method to identify and filter errors in the pseudo-depth map using multiple disparity maps by checking their consistency without the need for GT and a training process.
Experimental results show that the proposed method outperforms the previous methods and works well on various configurations by filtering out erroneous areas where the stereo-matching is vulnerable, especially such as textureless regions, occlusion boundaries, and reflective surfaces.
\end{abstract}
\vspace{-0.5em}
\section{Introduction}
\label{sec:intro}
\vspace{-0.5em}
Recently, monocular depth estimation has been studied since it plays a crucial role in robotics and autonomous driving for the perception of surroundings.
However, due to the high cost of acquiring ground truth (GT) for training, self-supervised learning framework \cite{garg2016unsupervised,godard2017unsupervised,godard2019digging} has been paid attention to as a solution to label-induced limitations.

Self-supervised learning framework has solved the limitations to some extent and can be divided into two groups: the photometric loss-based and stereo-matching knowledge distillation methods.
While the photometric loss-based methods \cite{godard2017unsupervised, godard2019digging} have used photometric loss as an alternative supervisory signal, stereo-matching knowledge distillation methods~\cite{guo2018learning, cho2019large, tonioni2019unsupervised, Choi_2021_ICCV} have used the self-supervised stereo-matching network to distill the knowledge to the monocular depth network through pseudo-depth maps obtained from the stereo-matching network.
Especially in the case of stereo-matching knowledge distillation methods, the methods have taken advantage of stereo-matching, which is more generalized even when trained only with synthetic datasets.
It is because stereo-matching learns the correspondence between the left and right images (disparity), while monocular depth estimation directly learns the depth from a single input image, a scale-ambiguous problem.

However, since pseudo-depth maps also can contain errors in the ill-posed regions, it is necessary to prevent the error from being transferred to the monocular depth network.
To deal with the error, the existing training frameworks include a learning-based stereo-confidence network~\cite{Poggi2016LearningFS,Kim_CVPR_2019}.
During training time, the confidence map is used as a weight map in loss computation to filter out the errors.
The previous methods~\cite{cho2019large, tonioni2019unsupervised, Choi_2021_ICCV} applied the fixed or learnable threshold to the confidence map.
Although the stereo-confidence networks have helped to improve depth performance, the networks need to be trained with GT and an extra training step.

In this paper, without GT and extra training for stereo-confidence, we propose a method to filter out errors in the pseudo-depth map provided by the stereo-matching network for the stereo-matching knowledge distillation-based monocular depth estimation.
To filter out errors, the proposed method generates a weight map by checking the consistency between multiple disparity maps instead of using the learning-based stereo-confidence network.
The multiple disparity maps are obtained from a stereo-matching network using the disparity plane sweep, a fundamental algorithm in stereo-matching. 
Experimental results show that the proposed method improves performance on different configurations of monocular depth networks, stereo-matching networks, and datasets.

\vspace{-0.5em}
\section{Related Work}
\label{sec:related}
\vspace{-0.5em}
The self-supervised monocular depth estimation methods can be divided into two groups: photometric loss-based and stereo-matching knowledge distillation methods.
Since the proposed method is built upon the stereo-matching knowledge distillation method, we briefly review the existing stereo-matching knowledge distillation methods.


In the existing stereo-matching knowledge distillation methods, the knowledge of the stereo-matching network is distilled into a monocular depth network through pseudo-depth maps generated by the pre-trained self-supervised stereo-matching network \cite{guo2018learning,cho2019large,tonioni2019unsupervised,Choi_2021_ICCV}.
However, the pseudo-depth maps contain errors due to the ill-posed regions.
Thus, previous methods focused on identifying unreliable pixels and preventing those pixels in training time.
Guo \textit{et al}. \cite{guo2018learning} utilized occlusion masks trained by left-right disparity consistency to fine-tune a stereo-matching network trained with a synthetic dataset on a real dataset.
Cho \textit{et al}. \cite{cho2019large} used stereo confidence network \cite{Poggi2016LearningFS} to estimate confidence map with fixed threshold.
Tonioni \textit{et al}. \cite{tonioni2019unsupervised} also used a stereo confidence network but with a learnable threshold.
Choi \textit{et al}. \cite{Choi_2021_ICCV} proposed ThresNet, which adjusts thresholds depending on the input confidence map.
Although the previous methods have shown remarkable performances, they commonly relied on a stereo-confidence network, which requires GT and a separate training process.
Unlike them, the proposed filtering method is not a learning-based method, thereby independent of the training process such for a stereo-confidence network.

\vspace{-0.5em}
\section{Proposed Method}
\label{sec:method}
\vspace{-0.5em}
Figure \ref{fig:depth_from_stereo} shows the principle of obtaining depth information from a stereo camera. 
To obtain the depth information, using a stereo camera, a pair of left and right images is captured by two calibrated cameras assuming a pinhole camera model. 
In the stereo camera, the distance between the two cameras is defined as baseline $B$, and the distance between the camera plane and image plane is defined as focal length $f$.
An object in the real world can be observed in each pixel coordinate of the left and right images.
The object is located in different positions of each pixel coordinate of the left and right images.
When overlapping two images, the distance in the pixel unit between each object in the left and right image coordinates is defined as the disparity $d$. 
With the known values of baseline $B$ and focal length $f$, if we know the disparity $d$, the depth $D$ between an object and a camera plane can be calculated by means of triangulation. 

\vspace{-0.5em}
\subsection{Preliminaries and Concept}
\begin{figure}[!t]
    \centering
    \includegraphics[width=0.95\columnwidth]{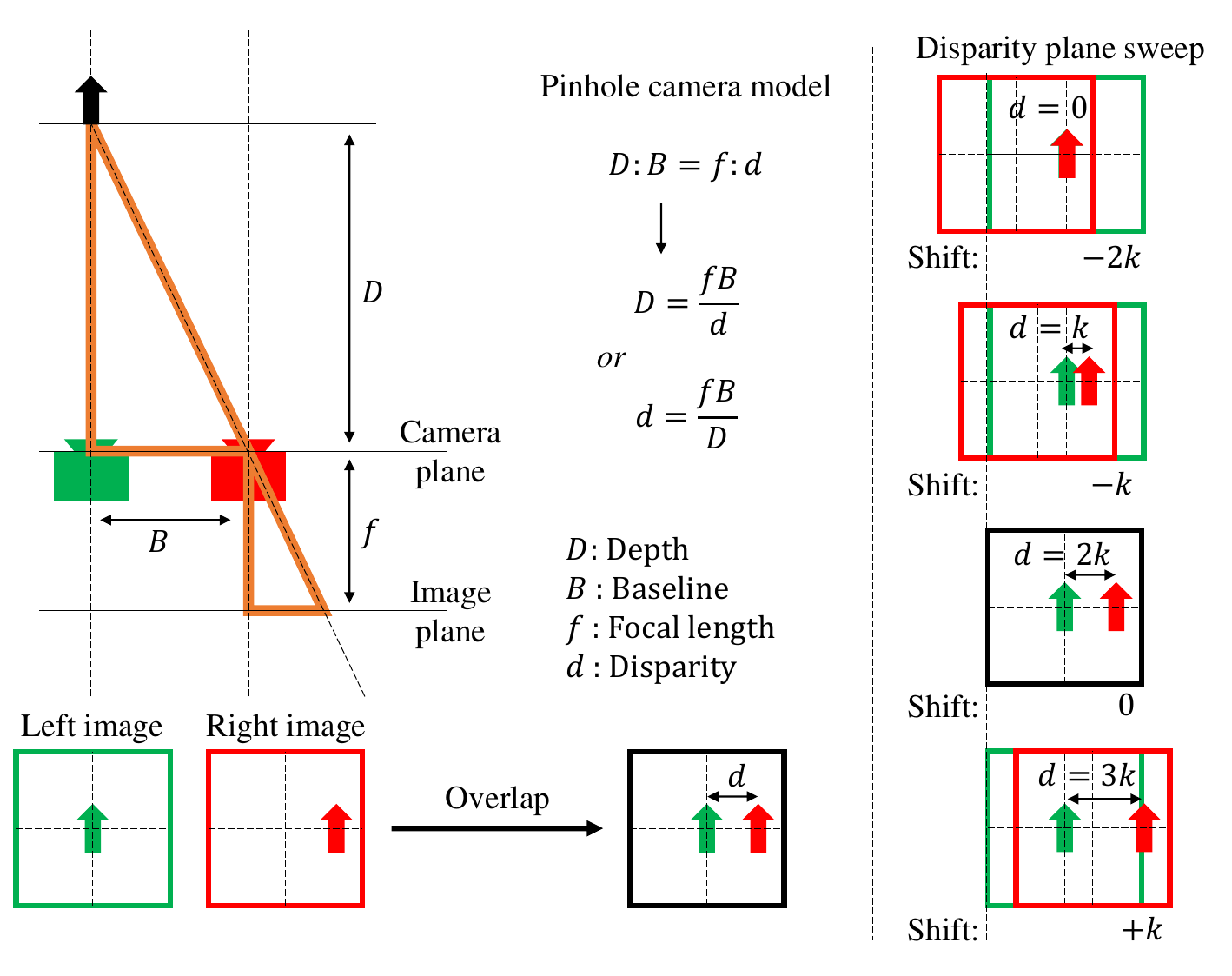}
    \caption{Principle to obtain depth from a stereo camera system. }
    \label{fig:depth_from_stereo}
\end{figure}

The stereo-matching aims to find a disparity value in the form of the 2-D image, i.e., the disparity map.
Generally, the disparity map is obtained with respect to the left image as a reference image.
By consecutively shifting the right image with respect to the left image, the stereo-matching methods calculate the intensity differences or similarity in pixel level between the left and right images, resulting in the cost volume. 
The disparity plane sweep is an algorithm consecutively shifting the right image with respect to the left image. 

Generally, in stereo-matching, the disparity plane sweep is utilized to construct cost volume, which is leveraged as one of the features.
However, unlike the conventional use of the disparity plane sweep, we utilize the disparity plane sweep to obtain multiple disparity maps to generate a weight map for filtering out unreliable pseudo-depth.
We focus on the definition of disparity and utilize a rule by the disparity plane sweep.
Let us explain the rule by the disparity plane sweep using an example (see right side of Fig. \ref{fig:depth_from_stereo}).
Assume that an object projected in the left and right images is overlapped by shifting $-2k$. 
When the right image is shifted $-2k$, the disparity of an object should be zero since the projected objects are completely overlapped.
On the other side, when the right image is shifted $0$, the disparity should be $2k$ because the displacement between objects in the left and right images is $2k$.
Using the multiple disparity maps generated through the disparity plane sweep, we check their consistency to see whether the rule is kept. 
If the correspondence of a point between the left and right images is clear (easy to find), the rule is more likely to be kept across the multiple disparity maps.
If not (hard to find), the rule is less likely to be kept.
If the rule is kept, the prediction can be considered reliable.
The rule was originally introduced in the light field (LF) \cite{FBS-SFA, LFUDM} with the relationship among the cost-based, foreground-background separation-based, and model-based methods according to the shift from a signal processing perspective.
Considering the connection between LF and stereo-matching, the proposed method brings the concept into stereo-matching and utilizes the rule to generate a weight map to filter out the unreliable pseudo-depth map obtained by the stereo-matching network.


\vspace{-0.5em}
\subsection{Overall Framework}
\begin{figure}[!t]
    \centering
    \includegraphics[width=0.95\columnwidth]{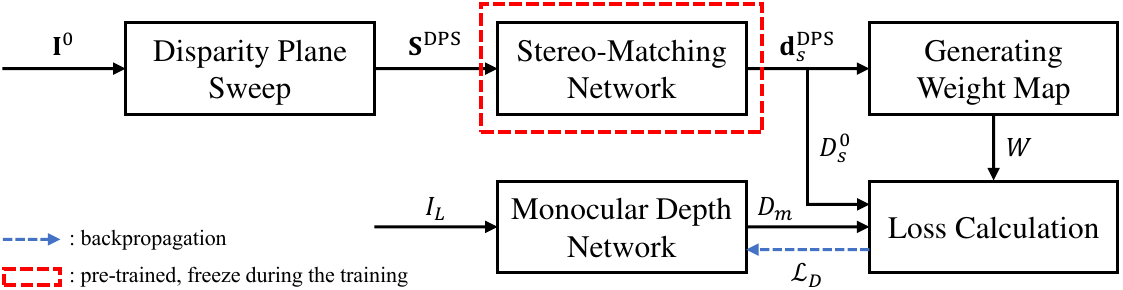}
    \caption{Overall framework of the proposed method.}
    \label{fig:proposed}
\end{figure}

\begin{figure}[!t]
    \centering
    \includegraphics[width=0.95\columnwidth]{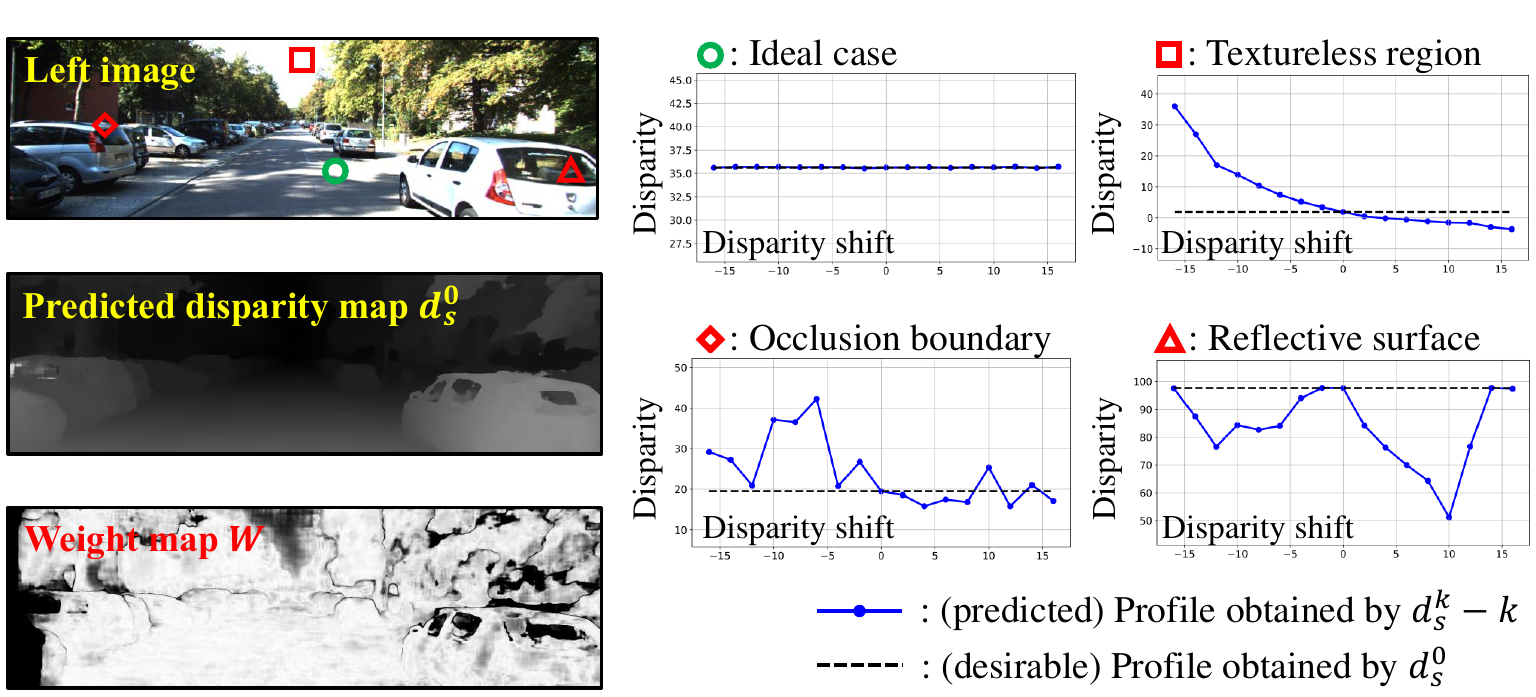}
    \caption{Disparity profile observation.}
    \label{fig:profile}
\end{figure}

Fig. \ref{fig:proposed} shows the overall framework of the proposed method. 
Built upon the stereo-matching knowledge distillation methods, we add the disparity plane sweep to obtain multiple disparity maps and replace the part of generating a weight map with the proposed method.
A set of disparity plane swept stereo image pairs $\mathbf{S}^{\text{DPS}}=\{\mathbf{I}^{-K},...,\mathbf{I}^{k},...,\mathbf{I}^{K}\}$ are generated using input stereo image pair $\mathbf{I}^{0}=\{I_{L}, I_{R}^{0}\}$, where $I_L$ and $I_R^{0}$ denote left and right images, respectively.
By shifting the right image by $k$ pixels between the disparity sweep range $k\in[-K,K]$, a disparity plane swept stereo image pair $\mathbf{I}^{k}=\{I_{L}, I_{R}^{k}\}$ is obtained.
Then, a set of disparity maps $\mathbf{d}_s^{\text{DPS}}=\{d_{s}^{-K}, ..., d_{s}^{k},..., d_{s}^{K}\}$ is acquired by feeding $\mathbf{S}^{\text{DPS}}$ to the pre-trained stereo-matching network~\cite{watson2020learning}, which is frozen during the training time.
Finally, a pixelwise weight map $W$ is obtained using $\mathbf{d}_s^{\text{DPS}}$. 
$W$ is used as a weight map in depth regression loss $\mathcal{L}_D$ between the predicted depth map of the monocular network $D_m$ and the pseudo-depth map $D_s^{0}$. 

\vspace{-0.5em}
\subsection{Generating Weight Map}
\vspace{-0.5em}
Since $d_{s}^{k}$ is obtained by the $k$ pixel shifted disparity plane swept stereo image pair $\mathbf{I}^{k}$, $d_{s}^{k}$ can be equal to $d_s^{0}$ in an ideal case by compensating $-k$, i.e., $d_s^{0}=d_{s}^{k}-k$.
Thus, at each pixel $p$, the unreliability score map $U(p)$ can be computed by 
\begin{equation}
    U(p) = \frac{1}{N-1}\sum_{k_n \in \mathbf{K}_N}\lVert d_{s}^{0}(p) - (d_{s}^{k_n}(p)-k_n) \rVert_{1},
\end{equation}
where $\mathbf{K}_N$ denotes a set of the $N$ number of disparity planes with constant step size and $k_n$ signifies its $n^{th}$ element. 
Using $U(p)$ and the maximum disparity $d_{max}$ of stereo-matching network, the weight map $W(p)$ can be obtained by 
\begin{equation} 
    W(p)=e^{-\sigma \frac{U(p)}{d_{max}}}, 
\end{equation}
where a scale factor $\sigma$ is set to have a $W(p)$ of 0.5 when $U(p)=1$.
Figure \ref{fig:profile} shows the proposed weight map and profiles of the ideal case and the ill-posed regions such as textureless region, occlusion boundary, and reflective surfaces.
For easy comparison, the dotted black lines represent $d_s^0$.
While the profile obtained by $d_s^k-k$ in the ideal case is shaped into a horizontal constant line as $d_s^0$, those in ill-posed regions are not due to matching ambiguity.
By filtering the ill-posed regions, a reliable pseudo-depth can be provided to monocular depth networks properly.

\vspace{-0.5em}
\subsection{Loss Function}
\vspace{-0.5em}
The disparity map without shifting ($d_s^{0}$) is converted to the pseudo depth map $D_s^{0}$ using focal length $f$ and baseline $B$, and the depth regression loss $\mathcal{L}_D$ is computed by
\begin{equation}\mathcal{L}_D=\frac{1}{Z}\sum_{p\in \Omega}W(p)\cdot\lVert D_{m}(p)-D_{s}^{0}(p) \rVert_{1},\end{equation}
where $\Omega$ denotes a set of pixel locations and $Z=\sum_{p\in \Omega}W(p)$.

\begin{table}[t!]
\centering
\caption{Quantitative evaluation on the KITTI Eigen split dataset. The best and second-best scores are bold-faced and underlined, respectively.}
\label{tab:comp}
\resizebox{\columnwidth}{!}
{
\begin{tabular}{crrrrrrrr}
\hline\hline
\multirow{2}{*}{Methods}					& \multicolumn{1}{c}{\multirow{2}{*}{\#param.}}	& \multicolumn{7}{c}{Metrics - $\downarrow$: lower is better, $\uparrow$: higher is better.}					\\ \cline{3-9}
											& 						& \multicolumn{1}{c}{Abs Rel $\downarrow$}	& \multicolumn{1}{c}{Sqr Rel $\downarrow$}	& \multicolumn{1}{c}{RMSE $\downarrow$}	& \multicolumn{1}{c}{RMSE log $\downarrow$}	& \multicolumn{1}{c}{$\delta_1 \uparrow$}	& \multicolumn{1}{c}{$\delta_2 \uparrow$}	& \multicolumn{1}{c}{$\delta_3 \uparrow$}	\\ \hline\hline
Monodepth \cite{godard2017unsupervised}		& 56M					& 0.138									& 1.186									& 5.65									& 0.234										& 0.813										& 0.930										& 0.969										\\
Monodepth2 \cite{godard2019digging}			& 14M					& 0.109									& 0.873									& 4.960									& 0.209										& 0.864										& 0.948										& 0.975										\\
Uncertainty \cite{Poggi_CVPR_2020}			& 14M					& 0.107									& 0.811									& 4.796									& 0.200										& 0.866										& 0.952										& 0.978                 					\\
MonoResMatch \cite{Tosi_2019_CVPR}			& 41M					& 0.111									& 0.867									& 4.714									& 0.199										& 0.864										& 0.954										& 0.979                 					\\
DepthHint \cite{watson2019self}				& 33M					& 0.102									& 0.762									& 4.602									& 0.189										& 0.880										& 0.960										& \underline{0.981}       					\\
Choi et al. \cite{Choi_2021_ICCV}			& 29M					& \underline{0.1}						& \textbf{0.644}						& \textbf{4.251}						& 0.187										& \underline{0.882}							& 0.960										& \underline{0.981}							\\ \hline
Baseline    								& 18M					& 0.106									& 0.803									& 4.478									& \underline{0.185}							& 0.881										& \underline{0.963}							& \textbf{0.983}							\\
Baseline + Ours								& 18M					& \textbf{0.098}						& \underline{0.650}						& \underline{4.316}						& \textbf{0.181}							& \textbf{0.890}							& \textbf{0.964}							& \textbf{0.983}							\\ \hline\hline \\[-1em]
\end{tabular}}
\end{table}

\begin{table}[t!]
\centering
\caption{Four sets of experiments with different configurations. For the backbone of the monocular depth network (Mono.), VGGNet16 \cite{simonyan2014very} and ResNet18 \cite{he2016deep} are used. For the stereo-matching networks (Stereo), Watson \textit{et al}.'s method \cite{watson2020learning} and Tonioni \cite{tonioni2019learning} are used to obtain pseudo-depth maps. For datasets, KITTI (K) and Cityscape (C) are used. For each set of experiments, the best scores are bold-faced. }
\label{tab:abl}
\resizebox{\columnwidth}{!}
{
\begin{tabular}{cccccccc}
\hline\hline
\multicolumn{4}{c}{Configurations} &  & \multicolumn{3}{c}{Metrics} \\ \cmidrule(rl){1-4} \cmidrule(rl){6-8}
Mono.       & Stereo & Dataset & Ours & &\multicolumn{1}{c}{Abs Rel $\downarrow$} & \multicolumn{1}{c}{RMSE $\downarrow$} & \multicolumn{1}{c}{$\delta_1 \uparrow$} \\ \hline\hline
ResNet18    & \cite{watson2020learning}   & K       & -                 && 0.113                        & 4.739                             & 0.867 \\
ResNet18    & \cite{watson2020learning}	  & K		& \checkmark              && \textbf{0.105}               & \textbf{4.555}					& \textbf{0.874} \\ \hline \\[-1em]
VGGNet16    & \cite{tonioni2019learning}  & K       & -                 && 0.103                        & 4.475								& 0.877 \\
VGGNet16    & \cite{tonioni2019learning}  & K		& \checkmark	            && \textbf{0.099}               & \textbf{4.415}	                & \textbf{0.882} \\ \hline \\[-1em]
VGGNet16    & \cite{watson2020learning}   & C	    & -				    && 0.104						& 5.708                             & 0.897 \\
VGGNet16	& \cite{watson2020learning}	  & C		& \checkmark	            && \textbf{0.101}               & \textbf{5.633}                    & \textbf{0.901} \\ \hline\hline \\[-1em]
\end{tabular}}
\end{table}

\begin{figure*}[!t]
    \centering
    \includegraphics[width=0.95\linewidth]{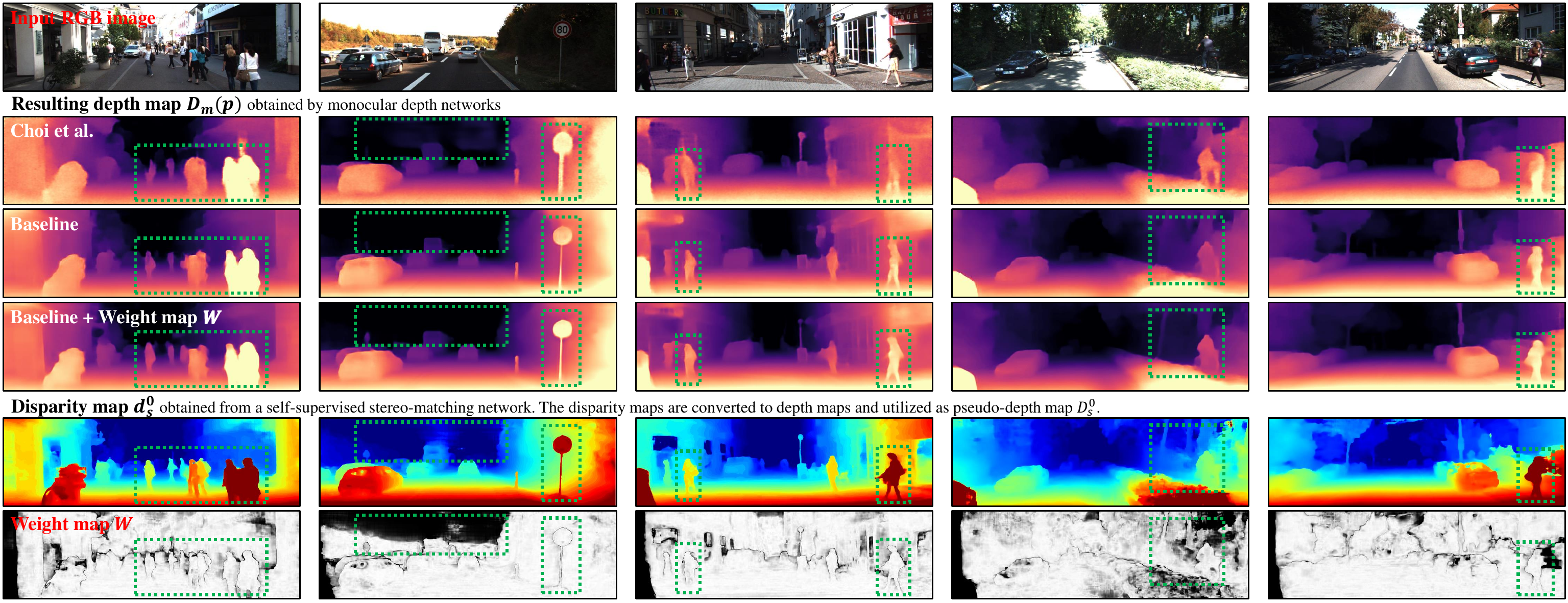}
    \caption{Qualitative comparison with the previous methods on the KITTI Eigen split dataset. $d_{s}^{0}$ and $W$ denote the predicted disparity map of the stereo-matching network \cite{watson2020learning} and the weight map of the proposed method (Ours), respectively.}
    \label{fig:qual_comp}
\end{figure*}

\vspace{-0.5em}
\section{Experiments}
\vspace{-0.5em}
\subsection{Experimental Setup}
\vspace{-0.5em}
Eigen split~\cite{eigen2014depth} of the KITTI~\cite{Geiger2013IJRR} dataset consisting of 24K images and the Cityscapes~\cite{Cordts2016Cityscapes} dataset consisting of 5K images are used to train and evaluate the proposed method.
We train the overall framework of the proposed method over 20 epochs with Adam optimizer and batch size of 12 at $192\times 640$ resolution.
The learning rate is set to $1e^{-4}$ for the first 15 epochs and reduced to $1e^{-5}$ for the remaining epochs.
In all experiments, $K$ and $N$ are set to 16 and 5, respectively.
For KITTI evaluation, we set the maximum depth value to 80 meters with Garg crop \cite{garg2016unsupervised}, and quantitative performance is evaluated by various metrics such as Abs Rel, Sqr Rel, RMSE, RMSE log, and accuracy within thresholds $\delta_1$, $\delta_2$, and $\delta_3$, which are introduced in Eigen \textit{et al}.'s method \cite{eigen2014depth}.
All experiments are conducted using PyTorch and a machine with 4 NVIDIA TITAN X GPUs.
For the baseline, we use U-Net architecture with VGGNet16 as a backbone network.
In all Tables, $\downarrow$ and $\uparrow$ denote that the lower and the higher are better, respectively.

\begin{table}[t!]
\centering
\caption{Comparison with other weight map generation methods on the KITTI Eigen split dataset.}
\label{tab:weight_map}
\resizebox{\columnwidth}{!}{
\begin{tabular}{cccccccc}
\hline\hline
\multirow{2}{*}{Weight map} & \multicolumn{7}{c}{Metrics} \\ \cline{2-8} 
                            & Abs Rel $\downarrow$  & Sqr Rel $\downarrow$  & RMSE $\downarrow$     & RMSE log $\downarrow$ & $\delta_1 \uparrow$   & $\delta_2 \uparrow$   & $\delta_3 \uparrow$   \\ \hline\hline
$\times$ (Baseline)                    & 0.106                 & 0.803                 & 4.478                 & 0.185                 & 0.881                 & 0.963                 & \textbf{0.983}        \\
CCNN \cite{CCNN}            & 0.101                 & 0.709                 & 4.295        & \textbf{0.180}        & 0.885                 & \textbf{0.964}        & \textbf{0.983}        \\ 
LGCNet \cite{LGC-Net}        & 0.104   & 0.707   & \textbf{4.250}   & 0.182   & 0.882   & \textbf{0.964}  & \textbf{0.983}   \\
Ours                        & \textbf{0.098}        & \textbf{0.650}        & 4.316                 & 0.181                 & \textbf{0.890}        & \textbf{0.964}        & \textbf{0.983}        \\
\hline\hline
\end{tabular}}
\end{table}

\vspace{-0.5em}
\subsection{Results}
In Table \ref{tab:comp}, we compare the baseline to which the proposed method is applied (Baseline + Ours) to the previous self-supervised methods using stereo pairs in training time \cite{godard2017unsupervised,godard2019digging,Poggi_CVPR_2020,Tosi_2019_CVPR,watson2019self} for the quantitative evaluation on the KITTI Eigen split.
Without the additional parameters, the Baseline + Ours achieves better performance except for Sqr Rel and RMSE, demonstrating its effectiveness and efficiency.
The Baseline + Ours also shows better qualitative results than the baseline (Baseline) and \cite{Choi_2021_ICCV} (see green dotted boxes in Fig. \ref{fig:qual_comp}).
The resulting depth maps of the Baseline + Ours show sharper results around object boundaries and maintain details, while the other methods often lose details.
The performance improvement over the baseline supports that the generated weight map (Ours) prevented the transfer of errors by reducing the weights of inaccurate regions in pseudo-depth maps during the training time.
In the last row of Fig. \ref{fig:qual_comp}, weight maps $W$ (Ours) for $d_{s}^{0}$ show that the unreliable regions are remarkably low.

Table \ref{tab:abl} shows experiments to validate the effectiveness of the proposed method with four configurations:
1) Backbone of the monocular depth network (Mono.), 2) Stereo-matching network (Stereo), 3) Dataset, and 4) Whether Ours is used or not. 
When the proposed method is used, the performances are consistently improved in different configurations.

In Table \ref{tab:weight_map}, we also compared the proposed method with CCNN \cite{CCNN} and LGCNet \cite{LGC-Net}, which are a learning-based stereo-confidence method, to check the effectiveness of the proposed method for generating a weight map.
The proposed method shows comparable performance to CCNN and LGCNet, even without extra training. 
It indicates that the proposed method effectively filters out the unreliable pseudo-depth.

\label{sec:typestyle}

\vspace{-0.5em}
\section{Conclusions}
\label{sec:majhead}
\vspace{-0.5em}
In this paper, we propose a method for filtering out unreliable pseudo-depth in the stereo-matching knowledge distillation methods. 
Based on the definition of disparity, using the multiple disparity maps, the proposed method effectively gives low weights to the incorrect regions in the pseudo-depth map.
Experimental result shows the potential of the proposed method through performance improvement without additional networks, GT and training steps.

\vspace{+0.5em}
\noindent\textbf{Acknowledgements.} This work was partly supported by Institute of Information \& communications Technology Planning \& Evaluation (IITP) grant funded by the Korea government(MSIT) (No.2021-0-02068, Artificial Intelligence Innovation Hub) and Artificial intelligence industrial convergence cluster development project funded by the Ministry of Science and ICT(MSIT, Korea)\&Gwangju Metropolitan City.
\vfill\pagebreak

\newpage
\bibliographystyle{IEEEbib}
\bibliography{strings,refs}

\end{document}